\documentclass[letterpaper]{article} 
\usepackage{aaai23}  
\usepackage{times}  
\usepackage{helvet}  
\usepackage{courier}  
\usepackage[hyphens]{url}  
\usepackage{graphicx} 
\usepackage{natbib}  
\usepackage{caption} 
\usepackage{amsmath}
\usepackage{algorithm}
\usepackage{algorithmic}
\usepackage{color,soul}
\usepackage{multirow}
\usepackage{bbold}

\usepackage{newfloat}
\usepackage{listings}
\DeclareCaptionStyle{ruled}{labelfont=normalfont,labelsep=colon,strut=off} 
\floatstyle{ruled}
\newfloat{listing}{tb}{lst}{}
\floatname{listing}{Listing}

\title{High-Throughput, High-Performance Deep Learning-Driven Light Guide Plate Surface Visual Quality Inspection Tailored for Real-World Manufacturing Environments}

\author{Carol Xu\equalcontrib\textsuperscript{\rm 1}, 
Mahmoud Famouri\equalcontrib\textsuperscript{\rm 1}, 
Gautam Bathla\equalcontrib\textsuperscript{\rm 1}, 
Mohammad Javad Shafiee\equalcontrib\textsuperscript{\rm 1,\rm 2},  Alexander Wong\equalcontrib\textsuperscript{\rm 1, \rm 2}
}

\affiliations{
    \textsuperscript{\rm 1}DarwinAI,  Waterloo, Ontario, Canada\\
    \textsuperscript{\rm 2}University of Waterloo, Waterloo, Ontario, Canada\\
    alexander.wong@uwaterloo.ca
}

\usepackage{bibentry}

\begin{document}

\maketitle

\begin{abstract}
Light guide plates are essential optical components widely used in a diverse range of applications ranging from medical lighting fixtures to back-lit TV displays.  An essential step in the manufacturing of light guide plates is the quality inspection of defects such as scratches, bright/dark spots, and impurities.  This is mainly done in industry through manual visual inspection for plate pattern irregularities, which is time-consuming and prone to human error and thus act as a significant barrier to high-throughput production.  Advances in deep learning-driven computer vision has led to the exploration of automated visual quality inspection of light guide plates to improve inspection consistency, accuracy, and efficiency.  However, given the computational constraints and high-throughput nature of real-world manufacturing environments, the widespread adoption of deep learning-driven visual inspection systems for inspecting light guide plates in real-world manufacturing environments has been greatly limited due to high computational requirements and integration challenges of existing deep learning approaches in research literature.  In this work, we introduce a fully-integrated, high-throughput, high-performance deep learning-driven workflow for light guide plate surface visual quality inspection (VQI) tailored for real-world manufacturing environments.  To enable automated VQI on the edge computing within the fully-integrated VQI system, a highly compact deep anti-aliased attention condenser neural network (which we name LightDefectNet) tailored specifically for light guide plate surface defect detection in resource-constrained scenarios was created via machine-driven design exploration with computational and ``best-practices'' constraints as well as L$_1$ paired classification discrepancy loss.  Experiments show that LightDetectNet achieves a detection accuracy of $\sim$98.2\% on the LGPSDD benchmark while having just 770K parameters ($\sim$33$\times$ and $\sim$6.9$\times$ lower than ResNet-50 and EfficientNet-B0, respectively) and $\sim$93M FLOPs ($\sim$88$\times$ and $\sim$8.4$\times$ lower than ResNet-50 and EfficientNet-B0, respectively) and $\sim$8.8$\times$ faster inference speed than EfficientNet-B0 on an embedded ARM processor.  As such, the proposed deep learning-driven workflow, integrated with the aforementioned LightDefectNet neural network, is highly suited for high-throughput, high-performance light plate surface VQI within real-world manufacturing environments.

\end{abstract}

\section{Introduction}
\label{sec:intro}

Deep neural networks have shown tremendous success in different applications and fields of research in the past decade. From computer vision tasks such as object detection~\cite{redmon2017yolo9000, liu2016ssd}, image segmentation~\cite{chen2017deeplab,he2017mask},  or object classification~\cite{krizhevsky2012imagenet,he2016deep} and anomaly detection~\cite{roth2022towards} to natural language processing problem like speech translation~\cite{vaswani2017attention}, question answering~\cite{ben2019block} or text to speech~\cite{graves2013speech}, these models exceeded the expectation of researchers  with unbelievable modeling accuracy. These encouraging results  have motivated different industries to take advantage of these models in their workflow and to introduce new automation processes for different tasks. 

However one of the bottlenecks of deep learning models is their computational complexity. These models are very computationally complex and parallel computing is one of the only ways to make them feasible to use. As such, adoption of these models has been more successful for the internet-based technologies because of the possibility of using cloud computing and taking advantage of powerful computing systems to make the process real-time. These models are still mostly infeasible to be used  in  off-line and remote environments where the could computing is not accessible.

Autonomous driving cars, surveillance cameras, manufacturing automation machines are some examples where cloud computation is not accessible and AI models should be processed offline and on edge devices. Limitation of processing power and/or memory in edge devices make the common deep neural network architectures  inefficient and  useless for these tasks which need to be performed in real-time. Visual quality inspection (VQI) in manufacturing automation is an example of this scenarios where if the VQI system does not perform in real-time it would reduce the throughput of the whole manufacturing pipeline and drop the efficiency dramatically. As such,  AI automation models must be highly efficient on edge devices.

\begin{figure*}[t]
    \centering
    \includegraphics[width=15cm]{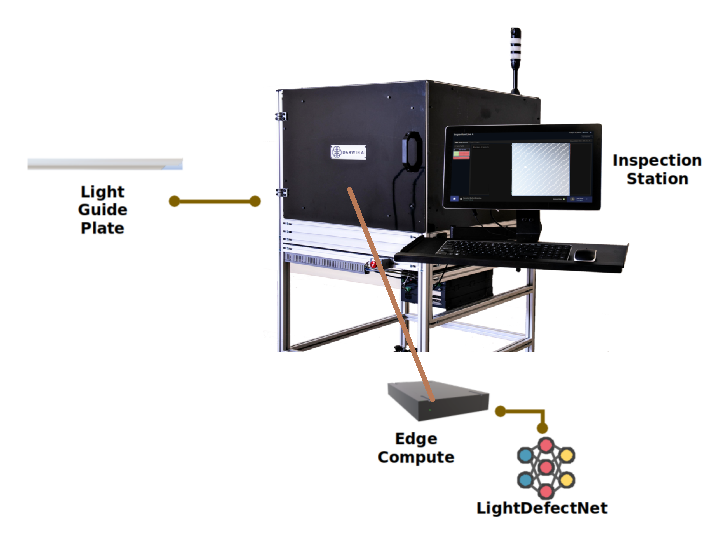}
    \caption{Proposed deep learning-driven light guide plate surface defect inspection workflow. A light guide plate on an assembly line is passed into a fully-integrated visual quality inspection (VQI) system.  The VQI system images the light guide plate, and the images are passed into an edge compute unit equipped with LightDefectNet to conduct high-speed surface defect detection.  The light on the VQI system would display different colors depending on whether a defect is detected on light guide plates being analyzed to alert an inspector.  An inspector may use the inspection station on the fully-integrated VQI system to examine the detected defective light guide plates. }
    \label{fig:mainflow}
\end{figure*}

Visual quality inspection is an important step in the manufacturing process where the manufactured parts need to be inspected and making sure they are defect-free, to be carried forward to the next step in the production line.  The conventional approach to perform this process has been to identify the possible defects by having human experts to monitor each part manually. However this approach is time-consuming and prone to human error due to repetitions of the tasks. On the other hand, the manual inspection can be a bottleneck in manufacturing throughput as human inspection efficiency bounded by  the human expert's  performance working in the line and limited. As such, automating this task has gained higher attention in the past decade.  Recently, the success of different AI algorithms has motivated industry to take advantage of AI model to address this problem. However, the hardware limitation in these environments make the adoption even more difficult. Small devices in this environments and limitation in power and memory budgets reduces the feasible computation hardware to edge devices. As such, edge AI and the ability to run AI model and specifically deep learning models on edge AI gain more traction recently.  

Edge AI and efficient deep learning models for edge devices has gained more attention in the past couple of years. Different pruning~\cite{li2016pruning}, or quantization~\cite{gholami2021survey} algorithms were proposed to reduce the number of parameters in the deep learning models or take advantage of hardware processing efficiency in quantized mode to speed up the running time for deep learning models. Moreover, automatic model optimization and AutoML~\cite{cai2019once,zhang2020fast,he2018amc} approaches have been the focused of the researchers to design efficient architectures in a more efficient and effective way. These types of algorithms  mostly formulate finding the efficient architecture as an optimization process satisfying an objective function. While these methods demonstrated a viable path to design efficient deep neural network architecture for edge devices they are still in their infancy and there is still not one solution to fit all.   

Moreover, these algorithms mostly have been evaluated by public datasets,  for example natural images in case of computer vision tasks; and when tried on specialized datasets, they fail to produce acceptable functioning accuracy and performance. Manufacturing automation tasks and specially visual inspection systems are one example.  

To this end, we proposed a new visual quality inspection model for light guide plates.    
Light guide plates~\cite{feng2004high} are one of the main optical components used in different devices such as medical lighting fixtures and both commercial flat panels and back-lit displays. In the LED LCD panels as an example, the light from the LED lamp passes through the light guide plate to be sprinkled on the surface of the large screen. The main role of the light guide plate is to  distribute the light evenly and to illuminate the surface. As such, it is very important to make sure there is no defect on the plate and visual inspection process plays a key role to identify possible defects.

Visual quality inspection of light guide plates is very challenging for a number of reasons;  the low contrast between the defect and the background, uneven brightness, and complex gradient texture make identifying the possible defects for an automated system tremendously difficult. As such, identifying possible defects on light guide plates are still  done mostly manually.

Recently the promises of deep learning~\cite{robinson2020learning,he2016deep,reiss2021panda,bergman2020classification} has motivated the development of new high performing deep neural networks for different manufacturing tasks for the purpose of improving the automation including the inspection systems~\cite{li2021end,shafiee2021tinydefectnet,9306797}. However, this field of research is still in its infancy given the constraints these types of systems require to satisfy and the limitations need to be addressed,  including i) high efficiency requirements, ii) with high accuracy and robustness, and iii) limited training data samples of different defected. Building deep neural networks satisfying the aforementioned constraints is time-consuming and usually impossible for non-expert users.

In this study, we take advantage of a machine-driven design exploration approach to specifically address the first two challenges and to a certain degree the third challenge of lacking enough training data by effectively exploring the architecture design space to automatically identify a tailored attention condenser network architecture (which we name LightDefectNet) based on computational and ``best-practices'' constraints. 

The main contributions of the manuscript is as follow:
\begin{itemize}
    \item A new visual quality inspection model for the light guide plates.  
    \item An efficient new neural network architecture designed automatically by taking advantage of Generative Synthesis framework. 
    \item A novel heterogeneous columnar design structure to account both for efficiency and effectiveness of feature representation in the designed deep neural network. 
    \item Illustrating the effective of anti-aliasing modules  compared to the traditional downsampling approach in deep learning structure.
    \item C omprehensive experimental results to evaluate the effectiveness of the proposed approach. 
\end{itemize}
In the next section we describe the designed architecture and the mythology followed by the comprehensive experimental results.

\section{Methodology}

\subsection{Deep Learning-Driven Light Guide Plate Surface Visual Quality Inspection Workflow}

The proposed deep learning-driven light guide plate surface defect inspection workflow is tailored specifically for real-world manufacturing environments and can be described as follows (see Figure~\ref{fig:mainflow}). During the light guide plate manufacturing process, the light guide plates on an assembly line is passed into a fully-integrated visual quality inspection (VQI) system.  The VQI system is capable of conducting high-speed captures of light guide plates to produce high-quality images to be used for automated analysis as well as parts quality tracking within a quality management system (QMS).  The captured images of the light guide plates are then passed into an edge compute unit equipped with LightDefectNet to conduct high-speed surface defect detection (the design of LightDefectNet will be described in detail in the next section).  The light on the VQI system would display different colors depending on whether a defect is detected on light guide plates being analyzed to alert an inspector, who may be tasked with multiple assembly lines at any given time and the visual alert will allow to quickly go to the appropriate line.  An inspector may then use the inspection station on the fully-integrated VQI system to examine the defective light guide plates detected using LightDefectNet to not only validate the results but also to remove the defective light guide plates from the line.

\subsection{Design of LightDefectNet via Machine-Driven Design Exploration}

In this study, we leverage the concept of Generative Synthesis~\cite{wong2018ferminets} to automatically identify the macro- and micro-architecture designs of the proposed LightDefectNet. 

\subsubsection{Generative Synthesis}
The Generative Synthesis framework (GS) is used to design and identify the efficient architecture for the visual inspection model. Generative Synthesis framework formulates the design exploration process  as a constrained optimization problem where the optimal network architecture is determined by  finding the optimal generator $G^\star(\cdot)$ which can generate network architectures $\{\mathcal{N}_s|s \in  S\}$ maximizing a universal performance function $U$ \cite{wong2019netscore} subject to a set of constraints:

\begin{align}
      G^{\star} = \underset{G^{'}} {\max} U \Big(G(s)\Big) \;\;\; \text{s.t.} \;\;\; \mathbb{1}_g(G(s)) = 1 \;\;\; \forall s \in S,
\end{align}
where $S$ is a set of seeds. The set of constraints are defined by a predefined set of operational requirements formulated via an indicator function $\mathbb{1}_g(\cdot)$.
The synthesis process is done within an iterative approach where at each step, the previous generator  $\bar{G}(\cdot)$ is evaluated by an inquisitor $I$ and based on its newly generated architectures $\mathcal{N}_s$. A new generator solution is evaluated based on the universal performance function $U$ by an indirect evaluation process.

More specifically, GS synthesizes new architectures by formulating the process as a  constrained optimization problem where the efficient network architecture is generated to satisfy different performance metrics such as accuracy and efficiency. The generated models are evaluated by a universal performance function which leads the optimization to the right path to reach a local optimal solution.   This constrained optimization is an iterative process and thus automatically determines the optimal architecture based on the provided requirements and input dataset, and therefore generates the optimal architecture that is tailored specifically for this application.

While GS is a generic neural architecture search method that can be leveraged for the purpose of machine-driven design exploration for any application, the operational requirements we enforce in the form of constraints (as follows) along with the input dataset for this specific application makes the search process itself customized specifically for this application.  Since this application in this study requires high throughput, high-accuracy inspection that runs on low-end edge compute hardware in the fully-integrated visual quality inspection (VQI) system, we specified the operational requirements around this to enable integrated deployment in real-world manufacturing environments.

\begin{figure*}[ht]
    \centering
    \includegraphics[width=13cm]{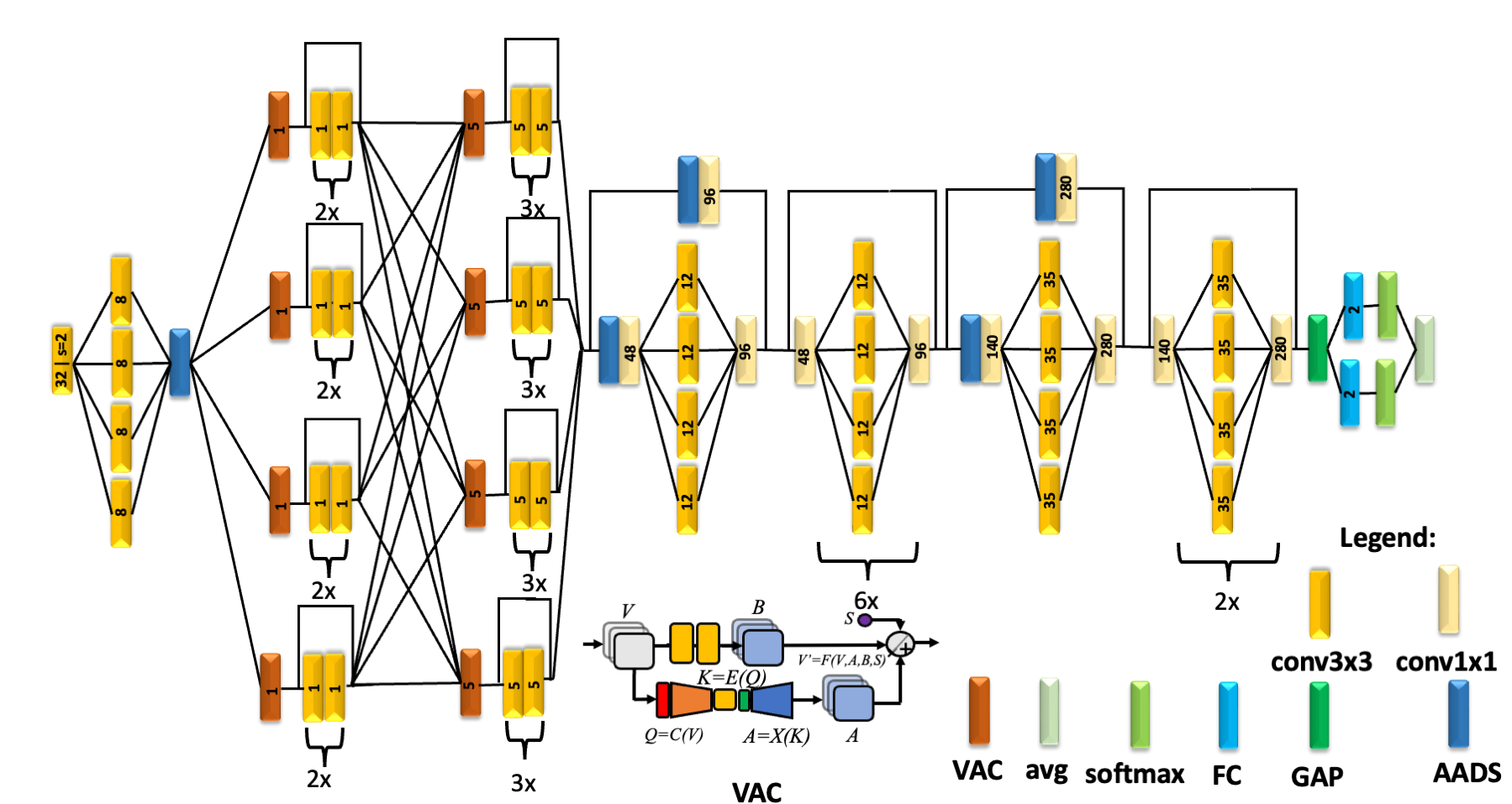}
    \caption{LightDefectNet architectural design;  the proposed attention condenser network architecture design is produced via a machine-driven design exploration, and possesses a heterogeneous columnar design, heterogeneous macroarchitecture and microarchitecture design diversity, early-stage self-attention via visual attention condensers (VAC), and anti-aliased downsampling (AADS) components for a great balance between detection accuracy, robustness, and complexity making it well-suited for this type of high-throughput and resource-constrained scenarios. Note that GAP stands for global average pooling.}
    \label{fig:architecture}
\end{figure*}


 We define a residual design prototype~\cite{he2016deep} to initialize the design process. The indicator function $\mathbb{1}_g(\cdot)$ is formulated such that it accounts for a combination of computational and ``best-practices'' constraints driven by architecture design lessons learned over the years by the community: i) number of floating-point operations (FLOPs) is under 100M FLOPs for resource-constrained manufacturing scenarios, (ii) pointwise strided convolutions (first introduced in the original residual network design~\cite{he2016deep} and continues to be leveraged in the recent RegNet design~\cite{radosavovic2020designing}) are restricted, as their use can lead to considerable information loss within the network, (iii) downsampling can only be conducted after the input layer via antialiasing downsampling (AADS)~\cite{zhang2019making}, as they have been shown to significantly improve network stability and robustness.  Furthermore, we incorporate attention condensers~\cite{wong2020tinyspeech, wong2020attendnets} as a viable design pattern into the machine-driven design exploration process, which are highly efficient self-attention mechanisms that learns and produces a condensed embedding characterizing joint local and cross-channel activation relationships for the purpose of selective attention. How and where attention condensers are leveraged, along with the rest of the micro-architecture and macro-architecture designs of LightDefectNet is left to the machine-driven design exploration process to automatically determine how best to satisfy the  constraints.

\subsubsection{Network Architecture Design}
 The network architecture design of LightDefectNet is shown in Figure~\ref{fig:architecture}. A number of key observations can be made about the generated LightDefectNet architecture; 
 \begin{enumerate}
     \item {\bf Early-stage self-attention:} it can be observed that visual attention condensers are leveraged heavily in the early stages of the network architecture.  The presence of visual attention condensers in these early stages enhance selective focus on important low-to-medium level visual indicators pertinent to light guide plate defects, while at the same time improving representational efficiency early on.
     \item \textbf{Heterogeneous columnar design:} a heterogeneous combination of columnar design patterns is exhibited in the proposed architecture design, with more independent columns with fewer stages of intermediate interaction in the earlier stages but more interactions in the later stages, which gives great balance between representational power and disentanglement with efficiency.  Furthermore, the fully-connected (FC) layer exhibits a columnar architecture where there are two softmax outputs which are then aggregated into the final softmax output to improve generalization and robustness particularly in low-data regimes.
     \item \textbf{Anti-aliased design:} a heavy use of anti-aliased downsampling (AADS) operations is exhibited across the architecture, which leads to improve robustness and stability compared to the use of traditional downsampling operations such as max-pooling. In this study, due to the columnar design of the FC layer, we leverage a loss function that maximizes the discrepancy between the soft outputs of the two FC columns to improve robustness and generalization of LightDefectNet in low-data regimes.
     \item \textbf{Heterogeneous design diversity:} heterogeneous microarchitecture and macroarchitecture design diversity is exhibited across the architecture due to the ability for the machine-driven design exploration strategy to determine the optimal microarchitecture designs in a fine-grained manner tailored around operational constraints and striking a strong balance between accuracy and efficiency. 
 \end{enumerate}

\section{Results \& Discussion}
The performance of the quality inspection framework is evaluated in regard to the designed LightDefectNet deep learning model used in the proposed framework.    
To explore the efficacy of the proposed LightDefectNet for light guide plates defect detection, we evaluated its performance on the LGPSDD (Light Guide Plate Surface Defect Detection) benchmark~\cite{9306797}. 

\begin{figure}
 \setlength{\tabcolsep}{0.1cm} 
\begin{tabular}{cc||cc}
     \multicolumn{2}{c}{\bf Defective}& \multicolumn{2}{c}{\bf Non-defective} \\ \hline \hline \\ 
     \includegraphics[width=2cm]{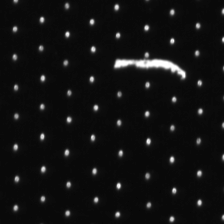}& \includegraphics[width=2cm]{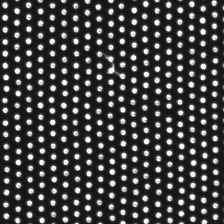}& 
     \includegraphics[width=2cm]{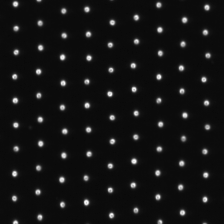}& \includegraphics[width=2cm]{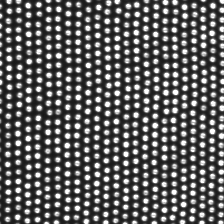} \\
     \includegraphics[width=2cm]{00209_NG_Image.png}& \includegraphics[width=2cm]{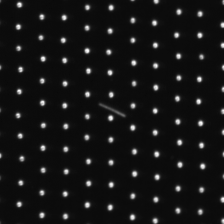}& 
     \includegraphics[width=2cm]{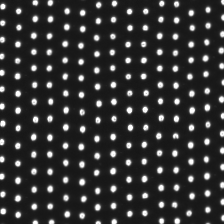}& \includegraphics[width=2cm]{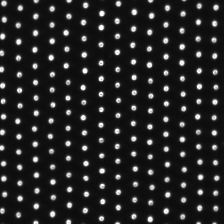} 
\end{tabular}
    \caption{Example of the light guide plates captured in the LGPSDD benchmark for both defective and non-defective samples.}
    \label{fig:data_example}
    \vspace{-0.5cm}
\end{figure}

\textbf{Benchmark data}: The LGPSDD (Light Guide Plate Surface Defect Detection) benchmark~\cite{9306797} used in this study consists of 822 images captured of light guide plates moving on a conveyor belt using an image acquisition platform that employs a line-scan camera and a multi-angle lighting source.  As shown in Figure~\ref{fig:data_example} the light guide plates captured in the LGPSDD benchmark has high physical diversity in terms of light guide point density, light guide brightness, as well as defect morphology and size~\cite{9306797}. This resulted in 422 defective samples and 400 non-defective samples, with a training/test split of 25\%/75\% as described in~\cite{9306797} to better mimic the typical low annotated data scenario seen in manufacturing applications.  The images are 224 $\times$ 224 in size, and these same dimensions were used in the input dimensions for the tested neural network architecture designs in this study.

The lack of samples is a major limitation for real-world manufacturing visual inspection scenarios, in particular in the light guide plate inspection application being tackled in this paper.  As such, this low-data challenge in real-world creation and deployment of machine learning for manufacturing visual inspection is very important to address.  The trend in the industry is to take advantage of off-the-shelf neural network architecture designs and train against a dataset, which would not work well in these types of real-world manufacturing applications.  As such, the limited training data samples that is leveraged as a part of the experimental setup and the practical considerations that we took into account in the design of the proposed system allows us to better understand the performance of the system in a more realistic manufacturing scenario.

\begin{center}
    \begin{table}
    \center
     
    \setlength{\tabcolsep}{0.1cm} 
    \begin{tabular}{l || r r r r }
        \multirow{ 2}{*}{ Model} &  Acc  &  Param & FLOPs &  Inf. Speed  \\
        &(\%) & (M) & (M) & (ms)\\
        \hline \hline
         ResNet-50 & 92.8 & 25.6 & 8200 & 83\\ 
        \hline 
         EfficientNet-B0 & 98.0 & 5.3 & 780 & 88\\
        \hline 
         MnasNet & 89.4 & 3.9 & 630 & 89\\ 
        \hline 
         MobileNetV3 (Large) & 97.8 & 5.4 & 438 & 56\\
        \hline
        LightDefectNet & \textbf{98.2} & \textbf{0.77} & \textbf{93} & \textbf{10}
    \end{tabular}
       \caption{ Quantitative results of the proposed LightDefectNet architecture compared to other tested architectures.}
    \label{tab:res}
    \end{table}
\end{center}

\textbf{Evaluated architectures}: In this study, in addition to the proposed LightDefectNet, we evaluated the performance of the ResNet-50~\cite{he2016deep} network architecture on the same LGPSSD benchmark for reference purposes, as well as several state-of-the-art efficient deep neural network architectures, including EfficientNet-B0~\cite{tan2020efficientnet}, MobileNetV3~\cite{howard2019mobilenet}, and MnasNet~\cite{tan2019MnasNet}. All architecture designs and experiments were conducted within the Pytorch deep learning framework.  In addition to quantitatively assessing the accuracy of each deep neural network architecture, we also evaluated the architectural complexity, theoretical computational complexity, as well as inference speed on an embedded ARM v8.2 64-bit 2.26GHz processor. 

\textbf{Training Policies}: All tested network architectures were trained using stochastic gradient descent optimization for 100 epochs and batch size of 5.  Different learning rates were used for optimal performance based on empirical analysis: $5.0\times10^{-5}$ for ResNet-50, $1.0\times10^{-3}$ for EfficientNet-B0 and LightDefectNet, $1.3 \times 10^{-3}$ for MobileNetV3, and $3.1\times10^{-4}$ for MnasNet.

\subsection{Results}

\textbf{Quantitative performance and complexity}: Table~\ref{tab:res} illustrates the quantitative performance, architectural complexity, computational complexity, and inference speed of the proposed LightDefectNet architecture compared to the ResNet-50 architecture as well as several state-of-the-art efficient architectures. A number of observations can be made from the quantitative results.  First of all, it can be observed that the proposed LightDefectNet architecture consists of just $\sim$770K parameters, which is significantly smaller than ResNet-50 as well as the competing state-of-the-art efficient architectures.  More specifically, LightDefectNet is $\sim$33$\times$ smaller compared to the ResNet-50 architecture while achieving higher accuracy, and $\sim$6.9$\times$ smaller compared to the highly efficient EfficientNet-B0 (the most accurate architecture outside of LightDefectNet). Second, in terms of computational complexity, the proposed LightDefectNet architecture requires only $\sim$93M FLOPs, which is significantly lower than ResNet-50 as well as the tested state-of-the-art efficient architectures.  More specifically, LightDefectNet requires $\sim$88$\times$ fewer FLOPs compared to the ResNet-50 architecture, and $\sim$8.4$\times$ fewer FLOPs compared to EfficientNet. Third, from an accuracy perspective, LightDefectNet achieved the highest accuracy amongst  the tested architectures.  These results illustrate the strong balance achieved by the proposed LightDefectNet in terms of accuracy, architectural complexity, and computational complexity, making it very well-suited for high-performance light guide plate defect detection in resource-constrained manufacturing environments.

\textbf{Embedded inference speed}: We further explore real-world operational efficiency of the proposed LightDefectNet architecture in embedded scenarios by evaluating its run-time latency (at a batch size of 10) on an embedded ARM v8.2 64-bit 2.26GHz processor and compared with the other tested architectures.  It can be observed from Table~\ref{tab:res} that the proposed LightDefectNet architecture achieves a runtime latency of 10~ms per sample, which is significantly lower than ResNet-50 as well as the tested state-of-the-art efficient architectures.  More specifically, LightDefectNet is 8.3$\times$ faster when compared to the ResNet-50 architecture, 8.8$\times$ faster when compared to the EfficientNet-B0 architecture, and 5.6$\times$ faster when compared to the MobileNetV3 architecture (the fastest architecture outside of LightDefectNet).  The significant speed advantages of the proposed LightDefectNet architecture when compared to the other tested architectures make it very well-suited for use on embedded edge compute devices for high-throughput manufacturing scenarios, as well as illustrate the effectiveness of utilizing a machine-driven design exploration strategy with ``best-practices'' constraints for creating highly customized network architectures tailored specifically for industrial tasks in manufacturing scenarios.

\textbf{Deployment:}
In terms of path to deployment, here we discuss the general workflow of a deployable VQI system for light guide plate inspection (realized around a real-world commercial VQI system available in the market) and how it is possible to integrate the generated efficient architecture based on the proposed machine-driven design exploration approach for detecting defects on light guide plates in a practical, high-throughput manner into this VQI system. Furthermore, the machine-driven design exploration approach used in  this study has been deployed for real-world operational scenarios and offered to users as a Generative Synthesis development platform from DarwinAI, and the resulting VQI system described in this study is an Automatic Mixed-Assembly Inspection (AMI) system from DarwinAI that is available in commercial form.

\section{Conclusions}

We proposed a new visual quality inspection framework for light guide plate surface inspection. We  utilize the machine-driven design exploration with computational and ``best-practices`` constraints as well as L$_1$ paired classification discrepancy loss for the creation of highly compact deep neural network architectures for the task of light guide plate surface defect detection. The proposed framework use the designed efficient model to inspect the samples in real-time for identifying the possible defect.  Experimental results demonstrated that the proposed LightDefectNet was able to achieve a detection accuracy of $\sim$98.2\% on the LGPSDD benchmark while possessing significantly reduced architectural and computational complexity when compared to state-of-the-art efficient deep neural network architectures.  Furthermore, we demonstrated that the proposed LightDefectNet achieves significantly faster inference speed on an embedded ARM processor, making it very well-suited for light guide plate defect detection in high-throughput, resource-constrained manufacturing scenarios.  As a future direction, we aim to further explore the leveraging of this machine-driven design exploration strategy for producing highly efficient yet high-performing deep neural network architectures for other critical manufacturing applications as well as for other sensing modalities such as acoustic sensors for predictive maintenance.

\bibliography{aaai23.bib}

\end{document}